\documentclass[twocolumn, switch]{article} 

\usepackage{preprint}

\usepackage{amsmath, amsthm, amssymb, amsfonts}

\usepackage[numbers,square]{natbib}
\usepackage[utf8]{inputenc}	
\usepackage[T1]{fontenc}	
\usepackage{xcolor}		
\usepackage[colorlinks = true,
            linkcolor = purple,
            urlcolor  = blue,
            citecolor = cyan,
            anchorcolor = black]{hyperref}	
\usepackage{booktabs} 		
\usepackage{nicefrac}		
\usepackage{microtype}		
\usepackage{lineno}		
\usepackage{float}			
\usepackage{subcaption}

\usepackage{lipsum}		

\usepackage{newfloat}
\DeclareFloatingEnvironment[name={Supplementary Figure}]{suppfigure}
\usepackage{sidecap}
\sidecaptionvpos{figure}{c}

\usepackage{titlesec}
\titlespacing\section{0pt}{12pt plus 3pt minus 3pt}{1pt plus 1pt minus 1pt}
\titlespacing\subsection{0pt}{10pt plus 3pt minus 3pt}{1pt plus 1pt minus 1pt}
\titlespacing\subsubsection{0pt}{8pt plus 3pt minus 3pt}{1pt plus 1pt minus 1pt}

\usepackage{tikz,xcolor,hyperref}

\definecolor{lime}{HTML}{A6CE39}
\DeclareRobustCommand{\orcidicon}{
	\begin{tikzpicture}
	\draw[lime, fill=lime] (0,0) 
	circle [radius=0.16] 
	node[white] {{\fontfamily{qag}\selectfont \tiny ID}};
	\draw[white, fill=white] (-0.0625,0.095) 
	circle [radius=0.007];
	\end{tikzpicture}
	\hspace{-2mm}
}
\foreach \x in {A, ..., Z}{\expandafter\xdef\csname orcid\x\endcsname{\noexpand\href{https://orcid.org/\csname orcidauthor\x\endcsname}
			{\noexpand\orcidicon}}
}

\title{CrowdSim2: an Open Synthetic Benchmark for Object Detectors}

\usepackage{xwatermark}
\newwatermark[firstpage,color=gray!60,angle=90,scale=0.32, xpos=3.9in,ypos=0]{\href{https://doi.org/10.5220/0011692500003417}{\color{gray}{Publication doi: 10.5220/0011692500003417}}}
\newwatermark[firstpage,color=gray!90,angle=0,scale=0.28, xpos=0in,ypos=-5in]{*correspondence: \texttt{mstaniszewski@polsl.pl}}

\usepackage{authblk}

\author[1]{Paweł Foszner\orcidA{}}
\author[1]{Agnieszka Szczęsna\orcidC{}}
\author[2]{Luca Ciampi\orcidD{}}
\author[2]{Nicola Messina\orcidD{}}
\author[3]{Adam Cygan}
\author[3]{Bartosz Bizoń}
\author[4]{Michał Cogiel\orcidE{}}
\author[4]{Dominik Golba\orcidF{}}
\author[5]{Elżbieta Macioszek\orcidG{}}
\author[1\thanks{}]{Michał Staniszewski\orcidH{}}

\affil[1]{Department of Computer Graphics, Vision and Digital Systems, Faculty of Automatic Control, Electronics and Computer Science, Silesian University of Technology, Gliwice, Poland; name.surname@polsl.pl}
\affil[2]{Institute of Information Science and Technologies, National Research Council, Pisa, Italy; name.surname@isti.cnr.it}
\affil[3]{QSystems.pro sp. z o.o. Mochnackiego 34, 41-907 Bytom, Poland; nsurname@qsystems.pro}
\affil[4]{Blees sp. z o.o. Zygmunta Starego 24a/10, 44-100 Gliwice, Poland; nsurname@blees.co}
\affil[5]{Department of Transport Systems, Traffic Engineering and Logistics, Faculty of Transport and Aviation Engineering, Silesian University of Technology, Katowice, Poland; name.surname@polsl.pl}

\begin{document}

\twocolumn[ 
  \begin{@twocolumnfalse} 
  
\maketitle

\begin{abstract}
Data scarcity has become one of the main obstacles to developing supervised models based on Artificial Intelligence in Computer Vision. Indeed, Deep Learning-based models systematically struggle when applied in new scenarios never seen during training and may not be adequately tested in non-ordinary yet crucial real-world situations. This paper presents and publicly releases \textit{CrowdSim2}, a new synthetic collection of images suitable for people and vehicle detection gathered from a simulator based on the \textit{Unity} graphical engine. It consists of thousands of images gathered from various synthetic scenarios resembling the real world, where we varied some factors of interest, such as the weather conditions and the number of objects in the scenes. The labels are automatically collected and consist of bounding boxes that precisely localize objects belonging to the two object classes, leaving out humans from the annotation pipeline. We exploited this new benchmark as a testing ground for some state-of-the-art detectors, showing that our simulated scenarios can be a valuable tool for measuring their performances in a controlled environment. 
\end{abstract}
\vspace{0.35cm}

  \end{@twocolumnfalse} 
] 

\section{Introduction}
\label{sec:introduction}

In recent years, Computer Vision swerved toward Deep Learning (DL)-based models that learn from vast amounts of annotated data during the supervised learning phase. These models achieved astonishing results in several tasks that nowadays are considered basic, such as image classification, causing interest 
in addressing more complex domains such as object detection \cite{Cafarelli_2022}, 
image segmentation \cite{Bolya_2019}, 
visual object counting \cite{counting_edge} \cite{video_counting} \cite{counting_cells}, people tracking \cite{tracking_staniszewski}, or even facial reconstruction \cite{peszor} and video violence detection \cite{s22218345}. However, these more cumbersome tasks often also require more structured datasets that come with challenges concerning bias, privacy, and cost in terms of human effort for the annotation procedure. 

Indeed, more complex tasks correspond to more elaborated labels, and for each data sample, the effort shifts from annotating an image to annotating the objects present in it, even at the pixel level. Furthermore, more challenging tasks often go hand in hand with more complex scenarios that may rarely occur in the real world, yet correctly handling them can be crucial. Finally, privacy concerns surrounding Artificial Intelligence-based models have become increasingly important, further complicating data collection. Consequently, labeled datasets are often limited, and data scarcity has become the main stumbling block for the development and the in-the-wild application of Computer Vision algorithms. Deep Learning-based algorithms systematically struggle in new scenarios never seen during the training phase and may not be adequately tested in non-ordinary yet crucial real-world situations.

One appealing solution that is recently arising relies on collecting \textit{synthetic data} gathered from \textit{virtual environments} resembling the real world. Here, by interacting with the graphical engine, it is possible to \textit{automatically} collect the labels associated with the objects of interest, cutting off the human effort from the annotation procedure, thus reducing the costs. Furthermore, these reality simulators provide frameworks where it is possible to create specific scenarios by controlling and explicitly varying the factors that characterize them. Hence, they represent the perfect environments where automatically acquiring labeled data for the training phase but also be used as controlled testing grounds for evaluating the performance capabilities of the employed models.

\begin{figure}
 \centering
  \includegraphics[trim=0 4cm 0 0, clip, width=0.95\linewidth]{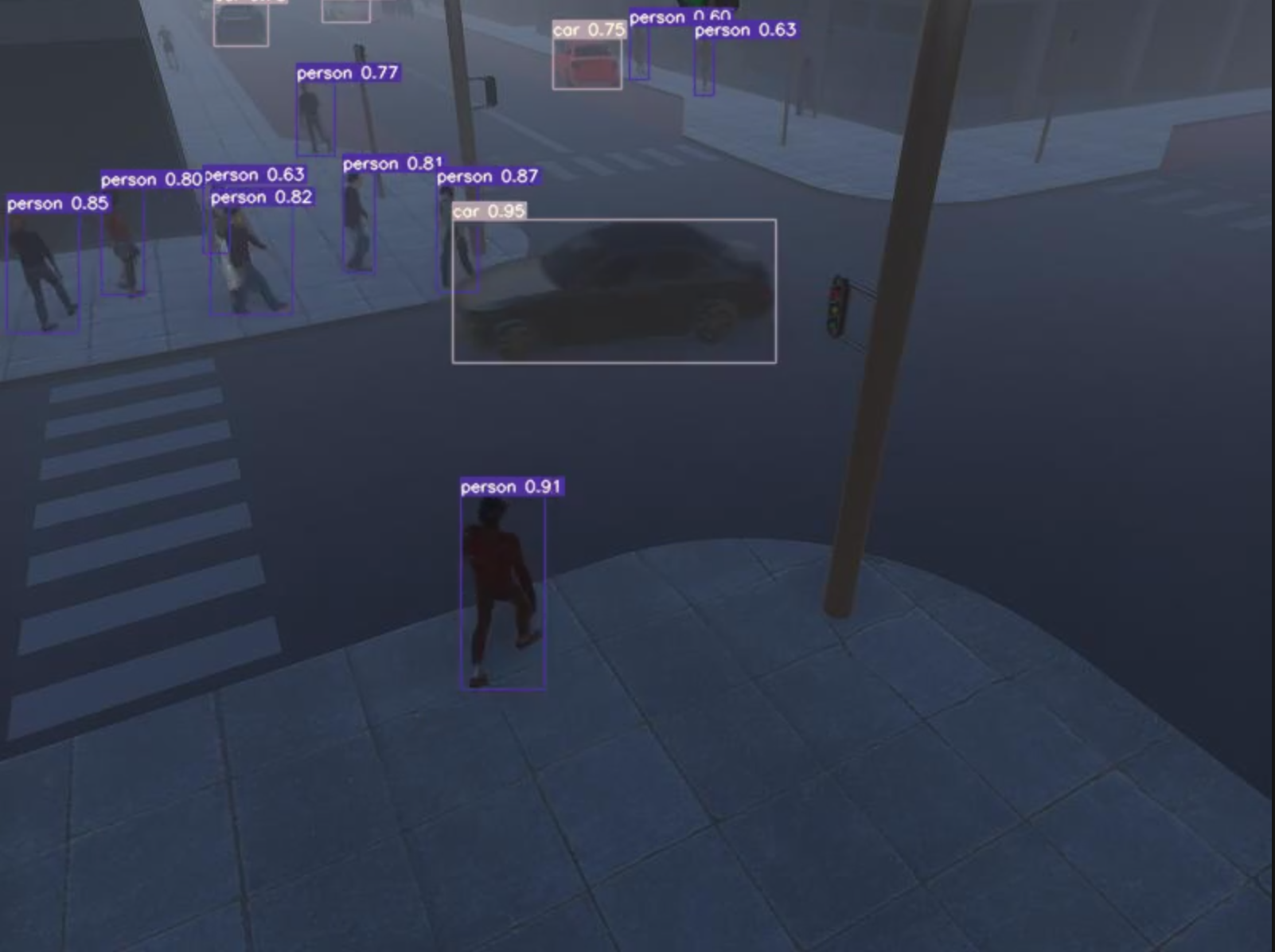}
  \includegraphics[trim=0 2.5cm 0 1.5cm, clip, width=0.95\linewidth]{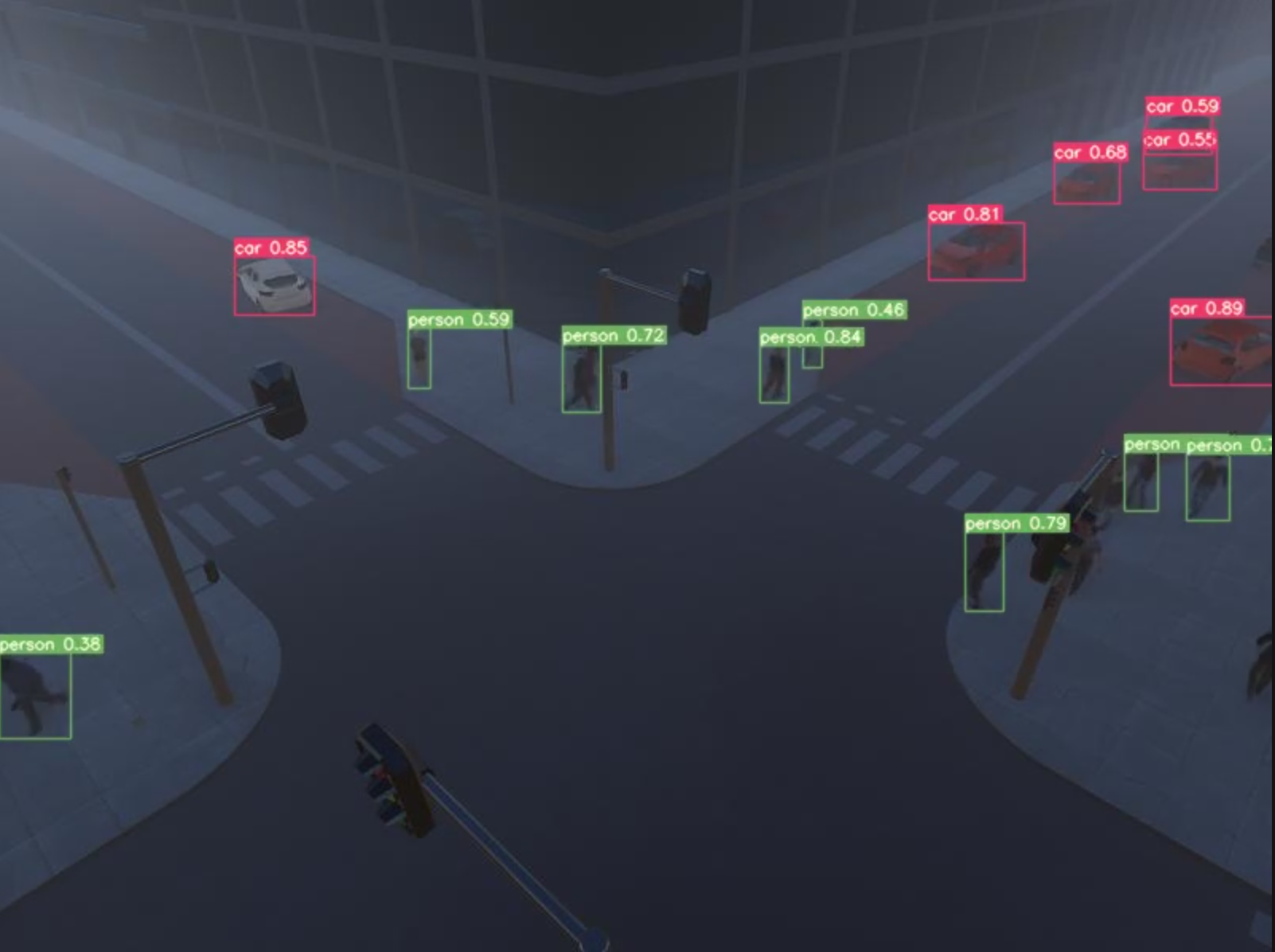}
   \vspace{0.2cm}
 \caption{Some samples of our synthetic dataset we rendered with our simulator, together with the bounding boxes localizing the objects of interest.}
 \label{fig:detections}
\end{figure}

In this paper, we consider the object detection task, focusing our attention on \textit{people} and \textit{vehicle} detection. We deem that people localization is crucial for security as well as for crowd analysis; on the other hand, vehicle detection constitutes the building block for urban and road planning, traffic light modeling, and traffic management, to name a few. In particular, we introduce and make publicly available \textit{CrowdSim2}, a new vast collection of synthetic images suitable for object detection and counting, collected by exploiting a simulator based on the \textit{Unity} graphical engine. Specifically, it consists of thousands of small video clips gathered from various synthetic scenarios where we varied some factors of interest, such as the weather conditions and the number of objects in the scenes. The labels are automatically collected and consist of bounding boxes that precisely localize objects belonging to two different classes --- \textit{person} and \textit{vehicle}. We report in Figure \ref{fig:detections} some samples of images together with the bounding boxes localizing the objects of interest in different scenarios we rendered with our simulator. Then, we present a detailed experimental analysis of the performance of several state-of-the-art DL-based object detectors pre-trained over general object detection databases present in the literature by exploiting our \textit{CrowdSim2} dataset as a testing ground. More in-depth, we extracted, from the collected videos, batches of frames belonging to specific and controlled scenarios, and we measured the obtained performances by varying the factors that characterized them.

Summarizing, the contributions of this paper are listed below:
\begin{itemize}
\item we propose \textit{CrowdSim2}, a new synthetic dataset suitable for \textit{people} and \textit{vehicle} detection, collected by exploiting a simulator based on the \textit{Unity} graphical engine and made freely available in the Zenodo Repository at \cite{crowdsim2};
\item we test some state-of-the-art object detectors over this new benchmark, exploiting it as a testing ground where we varied some factors of interest such as the weather conditions and the object density;
\item we show that our simulated scenarios can be a valuable tool for measuring detectors' performances in a controlled environment.
\end{itemize}

\section{Related Works}
\label{sec:related_works}

\subsection{Synthetic Datasets}
Synthetically-generated datasets have recently gained considerable interest due to the need for huge amounts of annotated data. Some notable examples are \textit{GTA5} \cite{Richter_2016} and 
\textit{SYNTHIA} \cite{Ros_2016} for semantic segmentation, \textit{Joint Track Auto (JTA)} \cite{jta} for pedestrian pose estimation and tracking, \textit{Virtual Pedestrian Dataset (ViPeD)} \cite{viped} \cite{viped_iciap} for pedestrian detection, \textit{Grand Traffic Auto (GTA)} \cite{da_visapp} for vehicle segmentation and counting, \textit{CrowdVisorPPE} \cite{Di_Benedetto_2022} for Personal Protective Equipment detection and \textit{Virtual World Fallen People (VWFP)} \cite{Carrara_2022} for fallen people detection. These datasets are mainly exploited for training deep learning models, which benefit from the fact that these collections of images are vast since the labels are automatically collected. On the other hand, using synthetic data as ground test collections is a relatively unexplored field. Furthermore, the datasets mentioned above are collected from the {GTA V (Grand Theft Auto V)} video game by {Rockstar North}. Although it is a very realistic generator of annotated images, some limitations arise when new scenarios or behaviors are needed. By contrast, using a simulator based on an open-source graphical engine allows one to create more customized environments and easily modify some factors of interest --- density of the objects, weather conditions, and object interactions.

\subsection{Object Detectors}
In the last decade, object detection has become one of the most critical and challenging branches of Computer Vision. It deals with detecting instances of semantic objects of a specific class (such as humans, buildings, or cars) in digital images and videos \cite{DBLP:journals/tcsv/DasiopoulouMKPS05}. 
This task has attracted increasing attention due to its wide range of applications and recent technological breakthroughs. Currently, most state-of-the-art object detectors employ Deep Learning models as their backbones and detection networks to extract features from images, classification, and localization, respectively. Existing object detectors can be divided into two categories: \textit{anchor-based} detectors and \textit{anchor-less} detectors. The models in the first category compute bounding box locations and class labels of object instances exploiting Deep Learning-based architectures that rely on anchors, i.e., prior bounding boxes with various scales and aspect ratios. They can be further divided into two groups: i) the two-stage paradigm, where a first module is responsible for generating a sparse set of object proposals and a second module is in charge of refining these predictions and classifying the objects; and ii) the one-stage approach that directly regresses to bounding boxes by sampling over regular and dense locations, skipping the region proposal stage. Some notable examples belonging to the first group are \textit{Faster R-CNN} \cite{fasterrcnn} and \textit{Mask R-CNN} \cite{mask_rcnn}. At the same time, popular networks of the latter set are the \textit{YOLO} family and \textit{RetinaNet} \cite{retinanet} algorithm.
On the other hand, anchor-free methods rely on predicting key points, such as corner or center points, instead of using anchor boxes and their inherent limitations. Some popular works existing in the literature are \textit{CenterNet} \cite{centernet}, and \textit{YOLOX} \cite{yolox}. 
Very recently, another object detector category is emerging, relying on the newly introduced Transformer attention modules in processing image feature maps, removing the need for hand-designed components like a non-maximum suppression procedure or anchor generation. Some examples are \textit{DEtection TRansformer (DETR)} \cite{detr} and one of its evolution, \textit{Deformable DETR} \cite{deformable_detr}.

In this paper, we consider some networks belonging to the \textit{"You Only Look Once" (YOLO)} family detectors, which turned out to be one of the most promising detector architectures in terms of efficiency and accuracy. The algorithm was introduced by \cite{yolo_v1} as a part of a custom framework called \textit{Darknet} \cite{redmon2013darknet}. Acronym \textit{YOLO (You Only Look Once)} derived from the idea of single shot regression approach. The author introduced the single-stage paradigm that made the model very fast and small, even possible to implement on edge devices.
The next version was \textit{YOLOv2} \cite{Redmon2017}, which introduced some iterative improvements (higher resolution, BatchNorm, and anchor boxes). \textit{YOLOv3} \cite{Redmon2018} added backbone network layers to the model and some other minor improvements. \textit{YOLOv4} \cite{yolo_v4} introduced improved feature aggregation and mish activation. \textit{YOLOv5} \cite{yolo5} proposed some improvements in feature detection, split into two stages - shallow feature detection and deep feature detection. The latest ones \textit{YOLOv6} \cite{yolo_v6} and \textit{YOLOv7} \cite{yolo_v7} added some new modules like the re-parameterized module and a dynamic label assignment strategy, further increasing the accuracy.

\section{The Crowdsim2 dataset}
\label{sec:dataset}

In this section, we introduce our \textit{CrowdSim2} dataset, a novel synthetic collection of images for \textit{people} and \textit{vehicle} detection \footnote{The dataset is freely available in the Zenodo Repository at \url{https://doi.org/10.5281/zenodo.7262220}}. First, we describe the Unity-based simulator we exploited for gathering the data, and then we depict the salient characteristics of this new database.

\subsection{The Simulator}
In this work, we exploited an extended version of the \textit{CrowdSim} simulator, introduced in \cite{staniszewski2020application}, that was designed and developed by using the \textit{Unity} graphical engine. The main goal of this simulator is to produce \textit{annotated} data to be used for training and testing Deep Learning-based models suitable for object and action detection. For this purpose, it allows users to generate realistic image sequences depicting scenes of urban life, where objects of interest are localized with precise bounding boxes. More in-depth, the simulator is designed using the \textit{agent-based} paradigm. In this approach, an agent -- in our work either a human or a vehicle -- is controlled individually, and decisions are made in the context of the environment in which the agent was placed. For instance, people can perform different types of movement thanks to the skeletal animation \cite{elsa} and actions depending on the situation in which they find themselves, including running, walking, jumping, waving or shaking hands, etc. The related animations vary depending on the age, height, and posture of the agent. Also, interactions between agents are possible in the so-called \textit{interaction zones}. Within this zone, the simulator continuously checks several conditions, such as the number of agents in the zone or random variables. If the conditions are met, the agents interact (fight, dance, etc.).

The environment in which agents are placed is important as the movement and behavior of the agents themselves. The considered simulator allows the user to generate a situation in four locations. They are:
\begin{itemize}
    \item traffic with intersections, pedestrian crossings, sidewalks, etc., in a typical urban environment, captured from three different cameras;
    \item a green park for pedestrians without traffic, filmed from three cameras;
    \item the main square of an old town, captured with two cameras;
    \item a tunnel for cars captured at both the endpoints, perfect for issues related to re-identification.
\end{itemize}

General rules of road traffic were applied to car movements. The starting positions of the cars are randomized among pre-defined starting points, and then the vehicles move to the point where they need to change direction. In such a place, cars make random decisions regarding further movements. Cars can only move in designated zones (streets and parking bays). 

\subsection{Simulated data}
\begin{figure*}[t]
 \centering
  \includegraphics[width=0.47\textwidth,height=3.8cm]{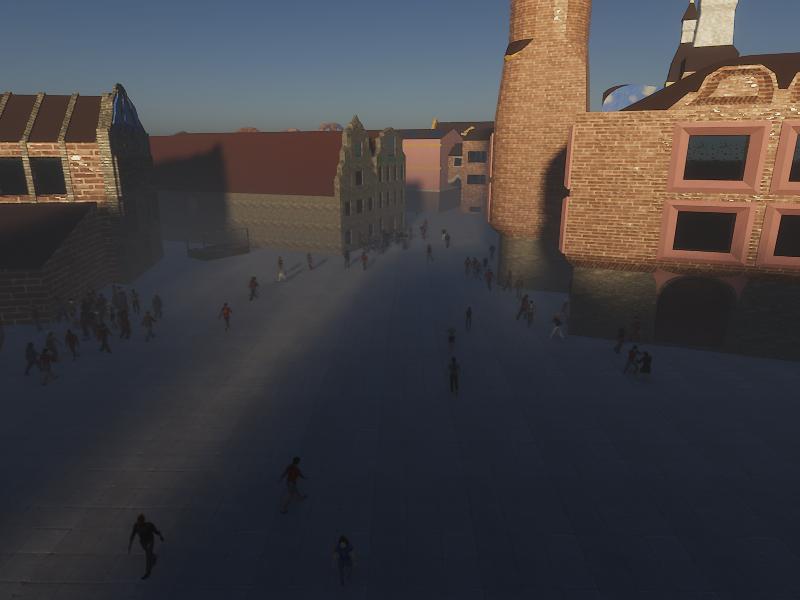}
  \includegraphics[width=0.47\textwidth,height=3.8cm]{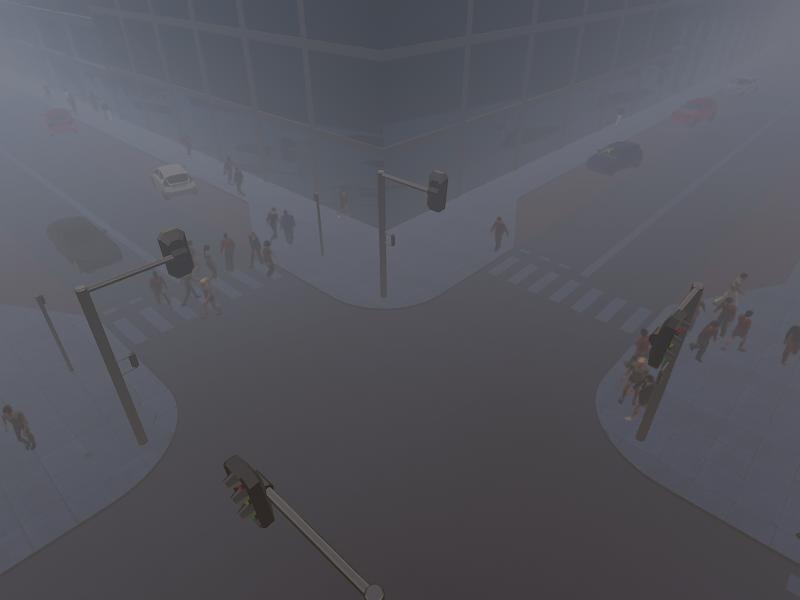} 
  \includegraphics[width=0.47\textwidth,height=3.8cm]{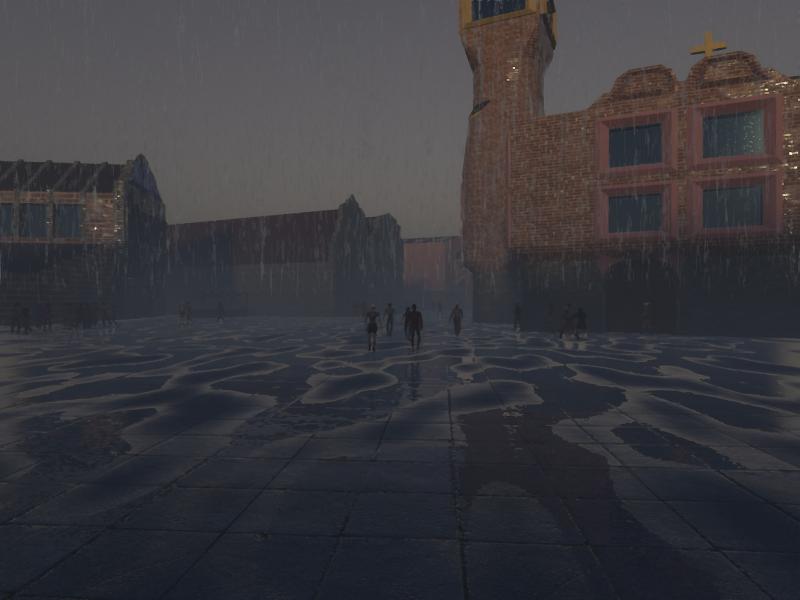}
  \includegraphics[width=0.47\textwidth,height=3.8cm]{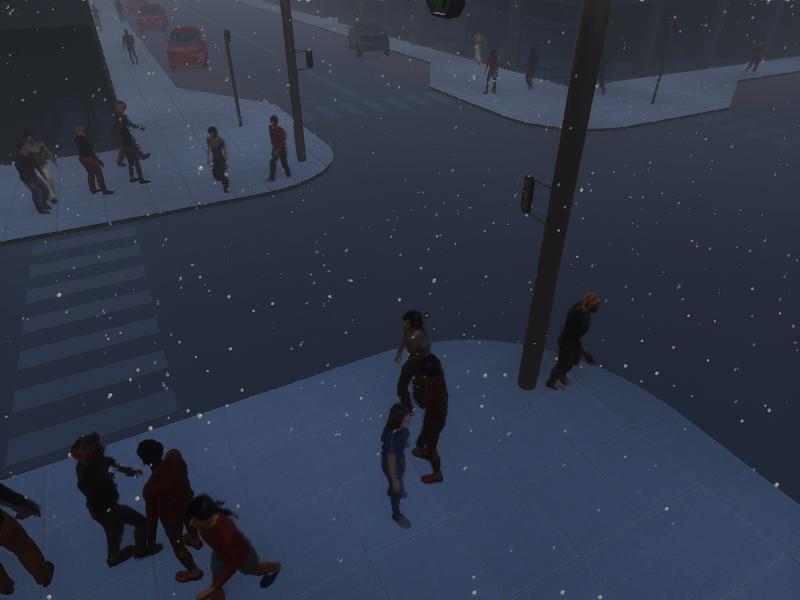}
   \vspace{0.2cm}
 \caption{Samples of our synthetic data where we show the four different weather conditions we varied with our simulator.}
 \label{fig:cond}
\end{figure*}

Using the simulator described in the previous section, we gathered a synthetic dataset suitable for \textit{people} and \textit{vehicle} detection. Specifically, for people detection, we used three different scenes, while for car detection, two different scenarios. We recorded thousands of small video clips of 30 seconds at a resolution of $800 \times 600$ pixels and a frame rate of 25 Frames Per Second (FPS), from which we extracted hundreds of thousands of still images. We varied several factors of interest, such as people's clothes, vehicle models, weather conditions (sun, fog, rain, and snow), and the objects' density in the scene. The ground truth is generated following the golden standard of the \textit{MOTDet Challenge} \footnote{\url{https://motchallenge.net/}}, consisting of the coordinates of the bounding boxes localizing the objects of interest --- \textit{people} and \textit{vehicles} in our case. The summary of the generated data is presented in Table \ref{tab:numbers}. We report in Figure \ref{fig:cond} the four different weather conditions we considered as one of the factors we varied during the data recording.

\begin{table}
\centering
\caption{Summary of our generated synthetic data. Each row corresponds to different weather conditions we set using our simulator. We report the total number of the collected video clips and the number of frames we extracted from them.}
\label{tab:numbers}
{%
\begin{tabular}{|c|c|c|}
\hline
& \# video-clips & \# frames \\ \hline
Sun  & 2,899                   & 2,174,250                \\ \hline
Rain & 1,633                         & 1,224,750                \\ \hline
Fog  & 1,653                     & 1,239,750               \\ \hline
Snow & 1,646                      & 1,234,500                \\ \hline
\end{tabular}
}
\end{table}

\begin{figure*}[t]
 \centering
 \begin{subfigure}[b]{0.24\textwidth}
  \includegraphics[trim=1.8cm 0 4.5cm 0, clip, width=\textwidth,height=2.8cm]{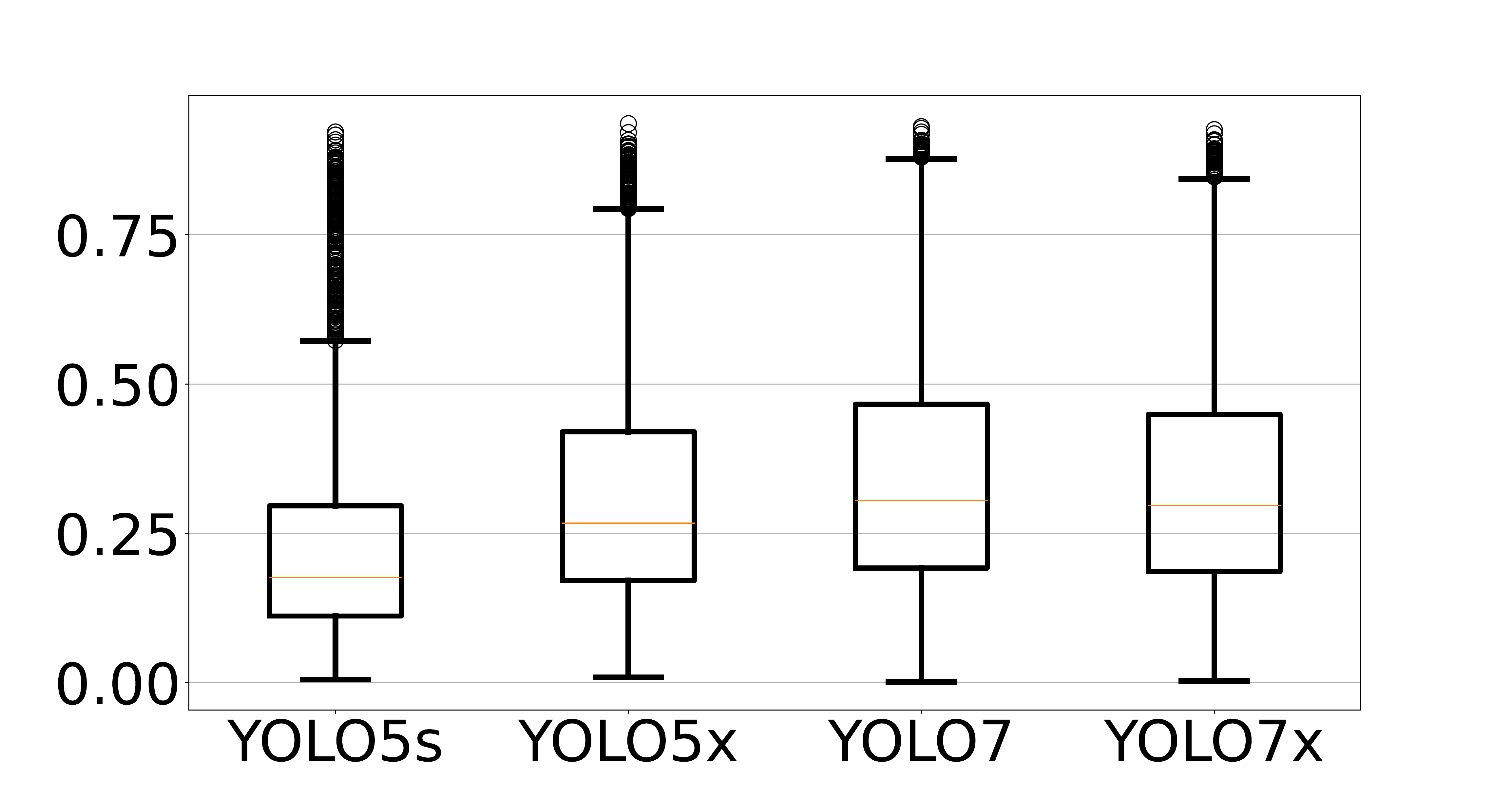}
  \caption{\textit{sun} weather condition.}
 \end{subfigure}
 \begin{subfigure}[b]{0.24\textwidth}
  \includegraphics[trim=1.8cm 0 4.5cm 0, clip, width=\textwidth,height=2.8cm]{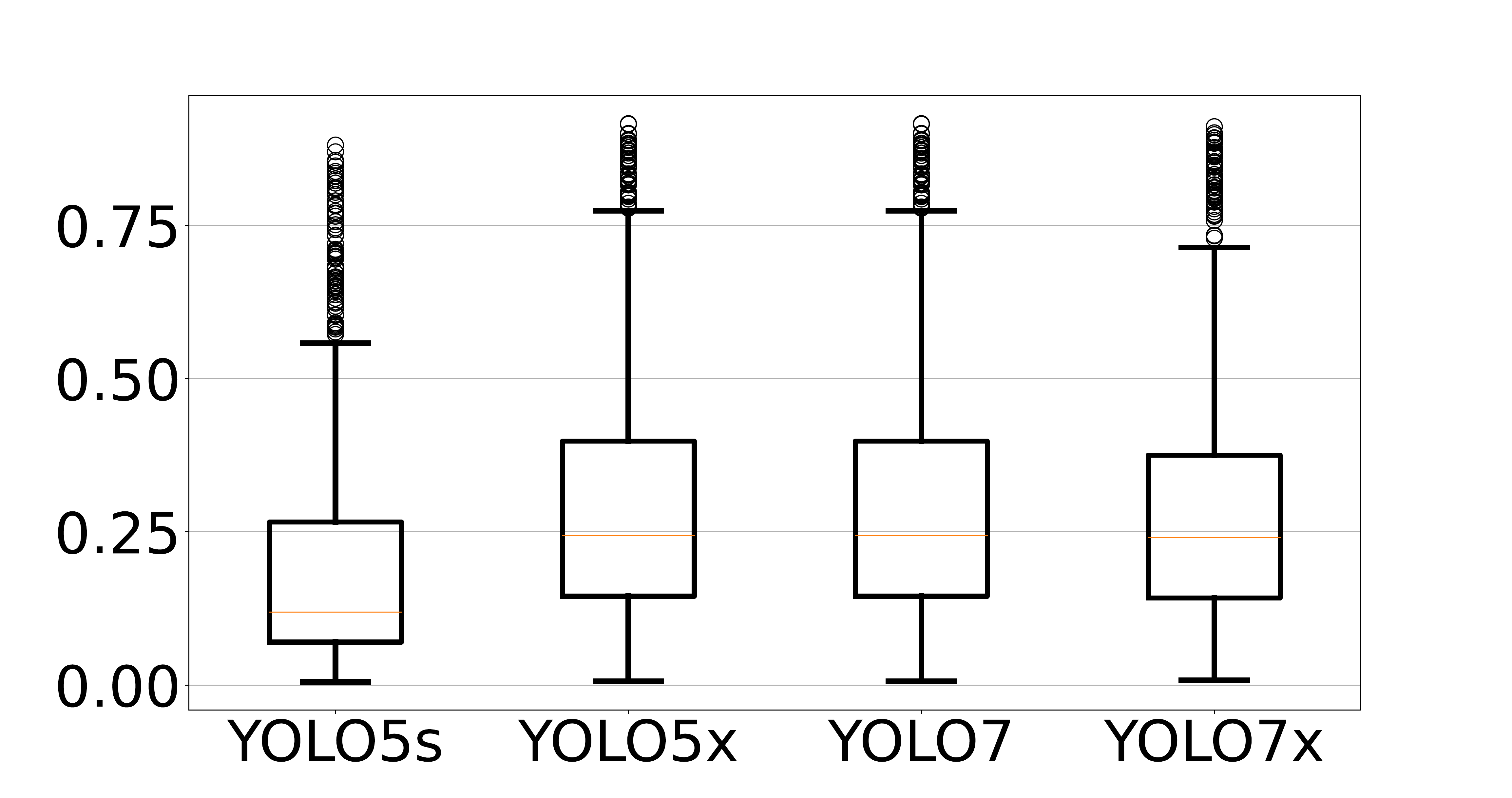}
  \caption{\textit{fog} weather condition.}
 \end{subfigure}
 \begin{subfigure}[b]{0.24\textwidth}
  \includegraphics[trim=1.8cm 0 4.5cm 0, clip, width=\textwidth,height=2.8cm]{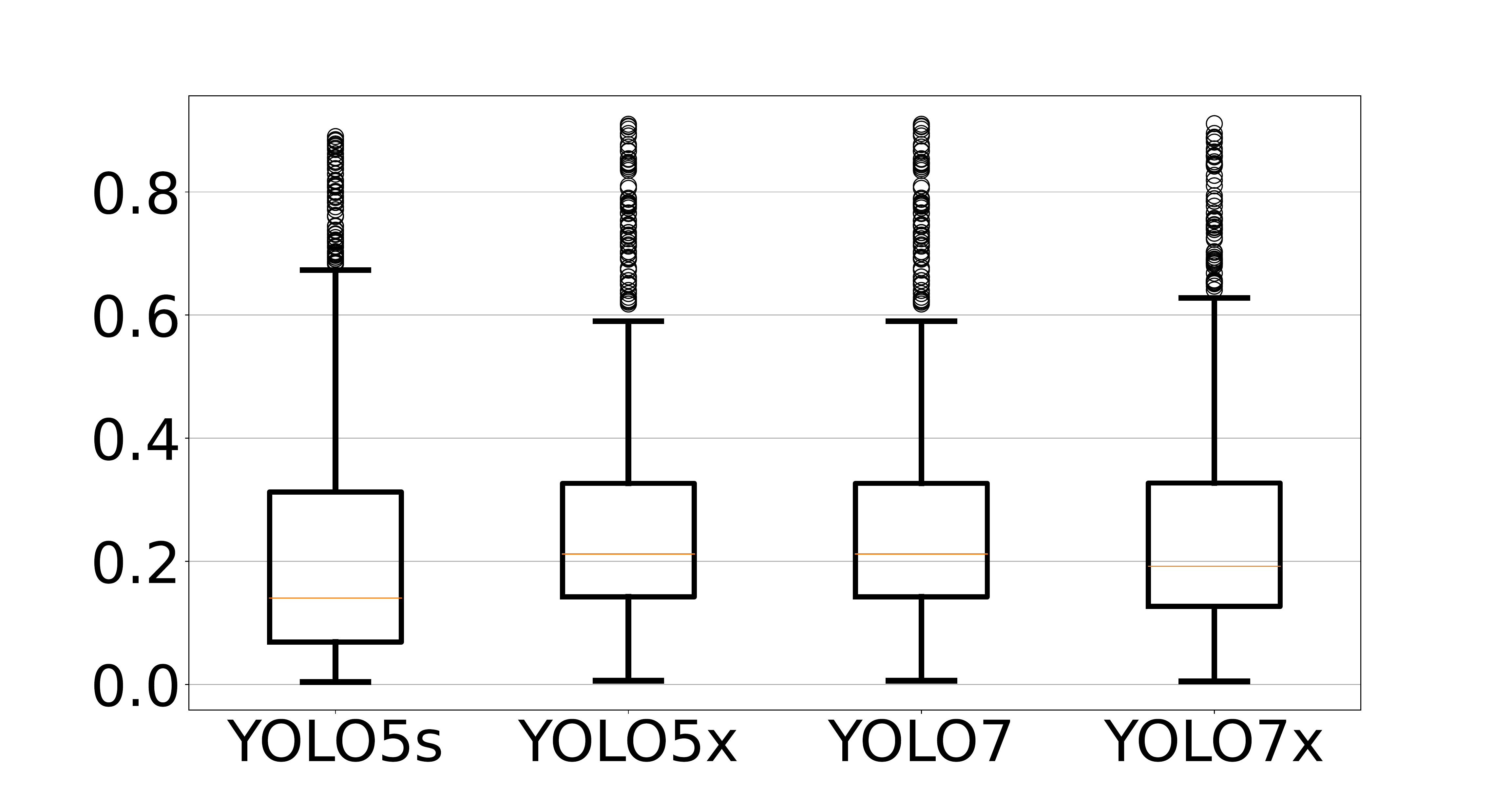}
  \caption{\textit{rain} weather condition.}
 \end{subfigure}
 \begin{subfigure}[b]{0.24\textwidth}
  \includegraphics[trim=1.8cm 0 4.5cm 0, clip, width=\textwidth,height=2.8cm]{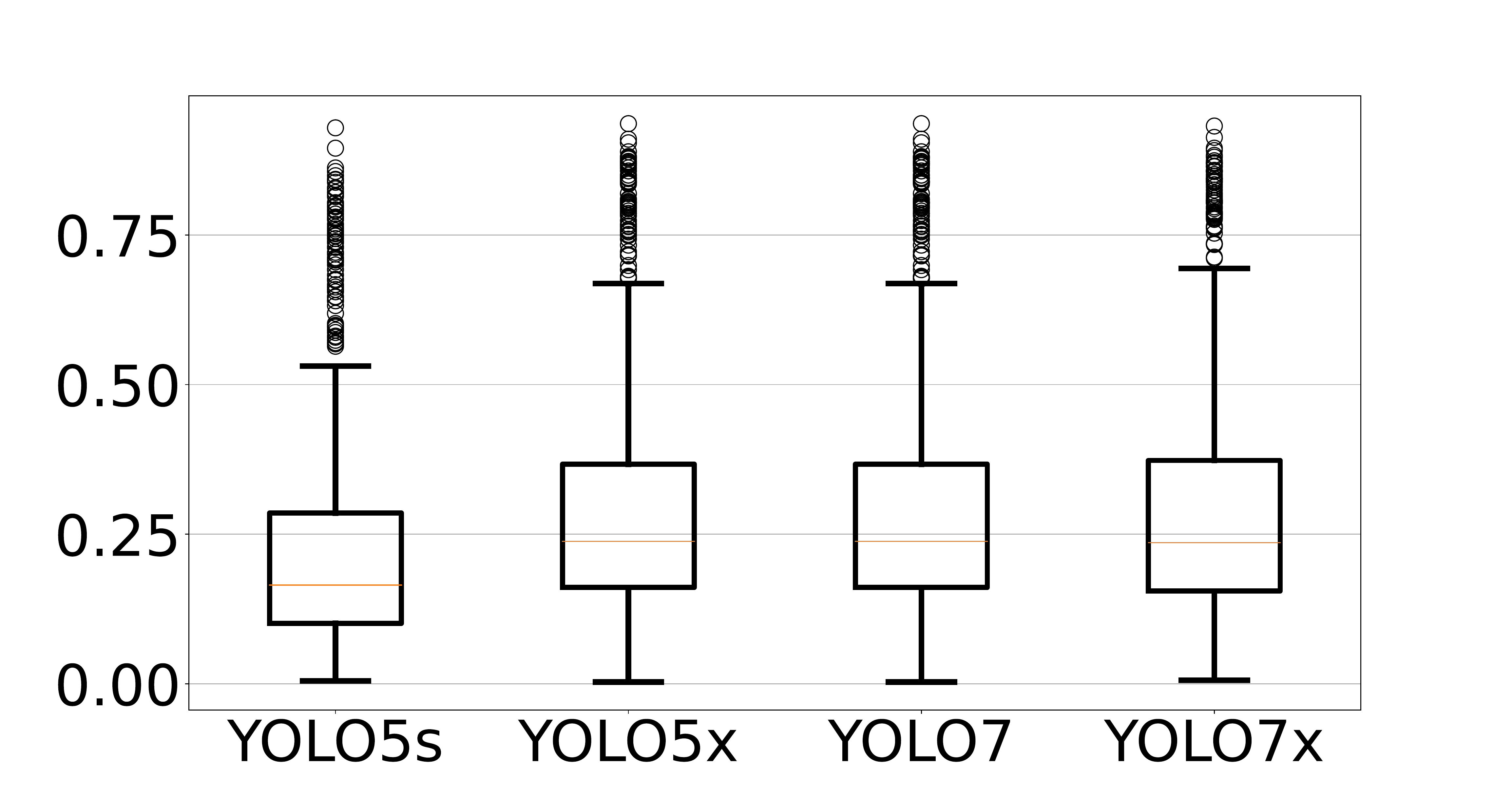}
  \caption{\textit{snow} weather condition.}
 \end{subfigure} 
 \vspace{0.2cm}
 \caption{Average Precision with IOU = 0.5  calculated for different weather conditions (\textit{sun, fog, rain and snow}), obtained for the \textit{people} detection task by exploiting the four considered \textit{YOLO} methods.}
 \label{fig:people_ap}
\end{figure*}

\begin{figure}
 \centering
  \begin{subfigure}[b]{0.4\textwidth}
    \includegraphics[width=\textwidth]{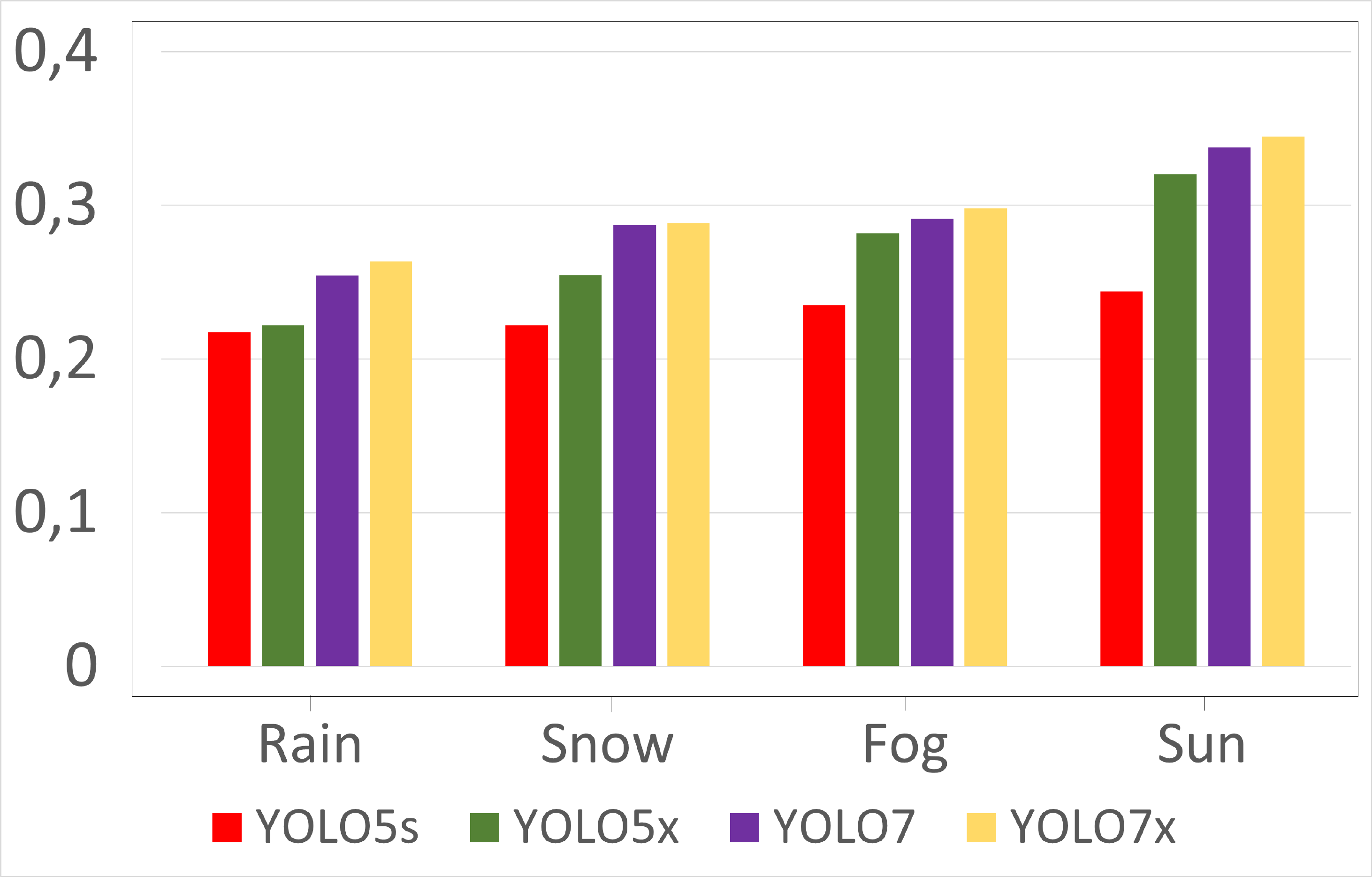}
    \caption{Results varying weather conditions.}
    \label{fig:people_sum_a}
  \end{subfigure}
    \begin{subfigure}[b]{0.42\textwidth}
    \includegraphics[width=\textwidth]{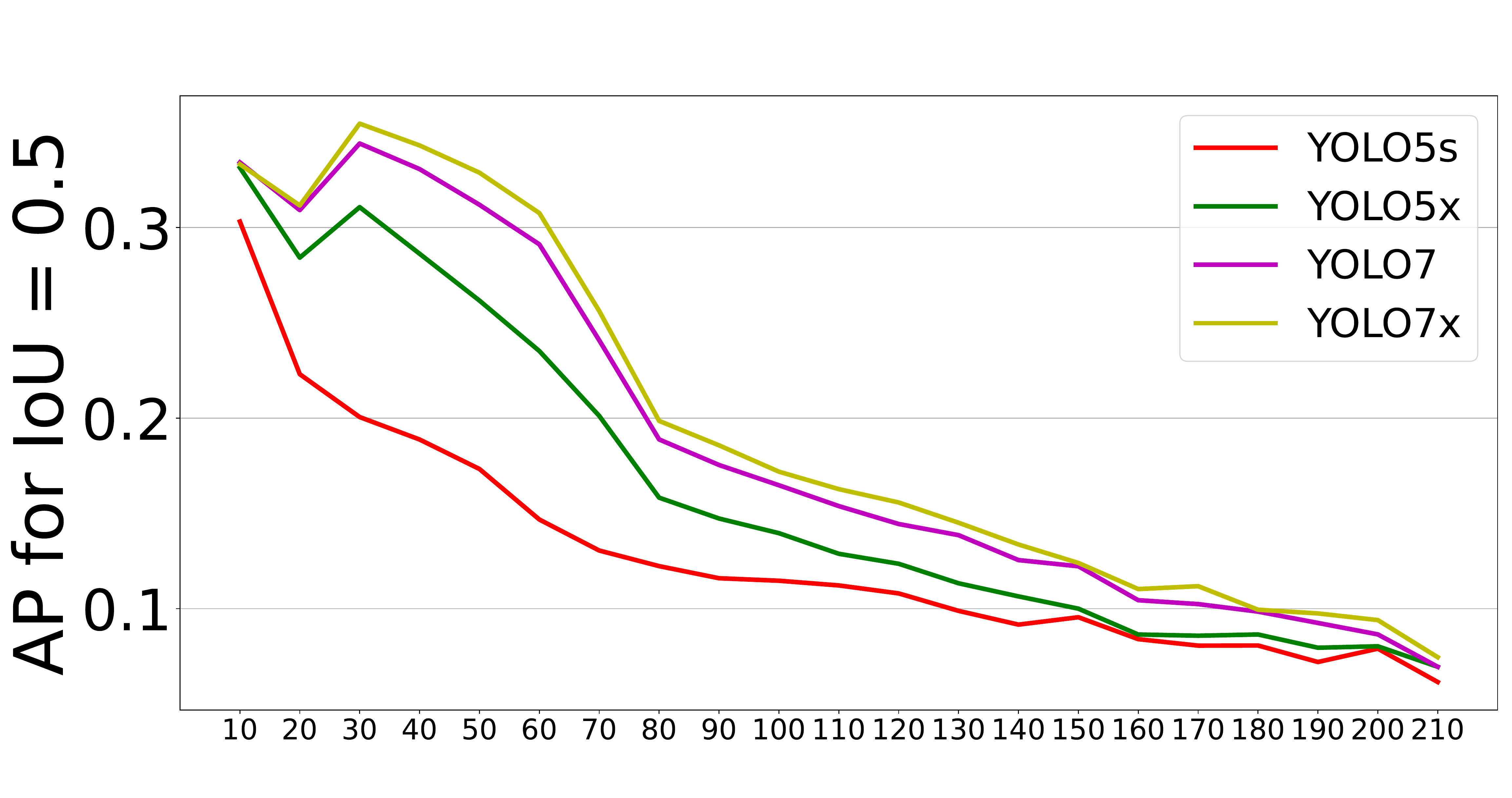}
  \caption{Results varying object densities.}
  \label{fig:people_sum_b}
  \end{subfigure}
  \vspace{0.2cm}
 \caption{Summary of Average Precision with IOU = 0.5 obtained with the four \textit{YOLO}-based considered methods by varying the two main simulated factors of interest: \textit{weather condition} and \textit{density} of the objects.}
 \label{fig:people_sum}
\end{figure}

\section{Results and discussion}
\label{sec:results}

In this section, we evaluate several deep learning-based object detectors belonging to the \textit{YOLO} family, described in Section \ref{sec:related_works}, on our \textit{CrowdSim2} dataset. Following the primary use case for this dataset explained in Section \ref{sec:introduction}, we employed it as a test benchmark to measure the performance of the considered methods in a simulated scenario where some factors of interest are controlled and changed. Specifically, we compared the obtained results considering four different weather conditions -- \textit{sun}, \textit{rain}, \textit{fog}, \textit{snow} -- and different densities of objects present in the scene -- from 1 object to hundreds of objects. 

We considered two different \textit{YOLO}-based models: \textit{YOLOv5} and \textit{YOLOv7}. Concerning \textit{YOLOv5}, we selected two different architectures having a different number of trainable parameters -- a light version we called \textit{YOLO5s} and a more deep architecture we referred to as \textit{YOLO5x}. Concerning \textit{YOLOv7}, we exploited the standard architecture (we referred to as \textit{YOLO7}) and a deeper version which we called \textit{YOLO7x}. Our decision to consider models having different architectures has been dictated by the fact that we wanted to prove that their behavior in the simulated data reflects the one observable over the real-world datasets -- shallow models are expected to exhibit moderate performances compared to deeper architectures. We refer the reader to Section \ref{sec:related_works} and the related papers for further details about the architectures of the employed detectors. All the models were fed with images of $640\times640$ pixels, and the models were pre-trained using the \textit{COCO} dataset \cite{coco}, a popular collection of images for general object detection. 

\begin{figure*}
 \centering
 \begin{subfigure}[b]{0.24\textwidth}
  \includegraphics[trim=1.8cm 0 4.5cm 0, clip, width=\textwidth,height=2.8cm]{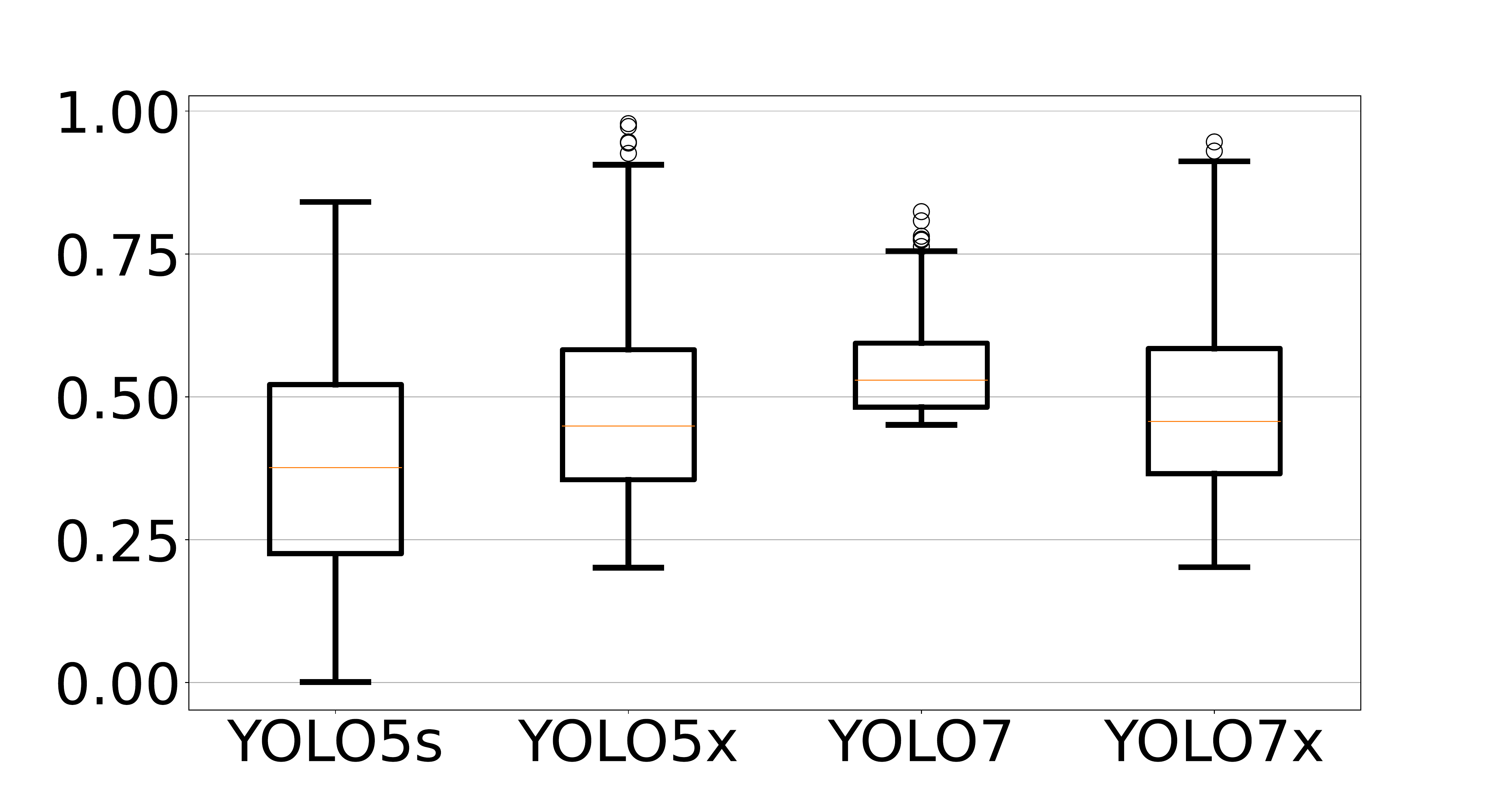}
  \caption{\textit{sun} weather condition.}
 \end{subfigure}
 \begin{subfigure}[b]{0.24\textwidth}
  \includegraphics[trim=1.8cm 0 4.5cm 0, clip, width=\textwidth,height=2.8cm]{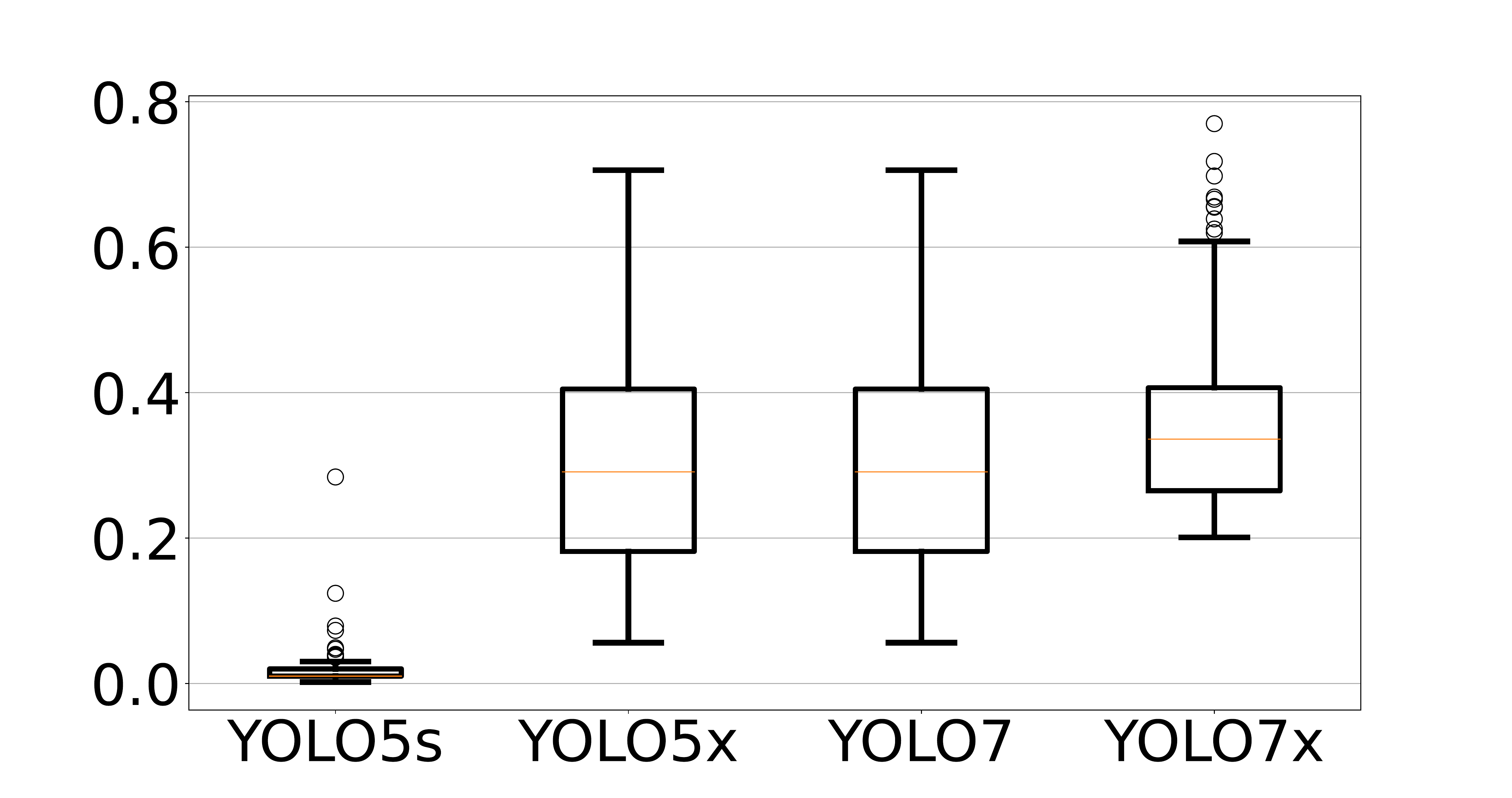}
  \caption{\textit{fog} weather condition.}
 \end{subfigure}
 \begin{subfigure}[b]{0.24\textwidth}
  \includegraphics[trim=1.8cm 0 4.5cm 0, clip, width=\textwidth,height=2.8cm]{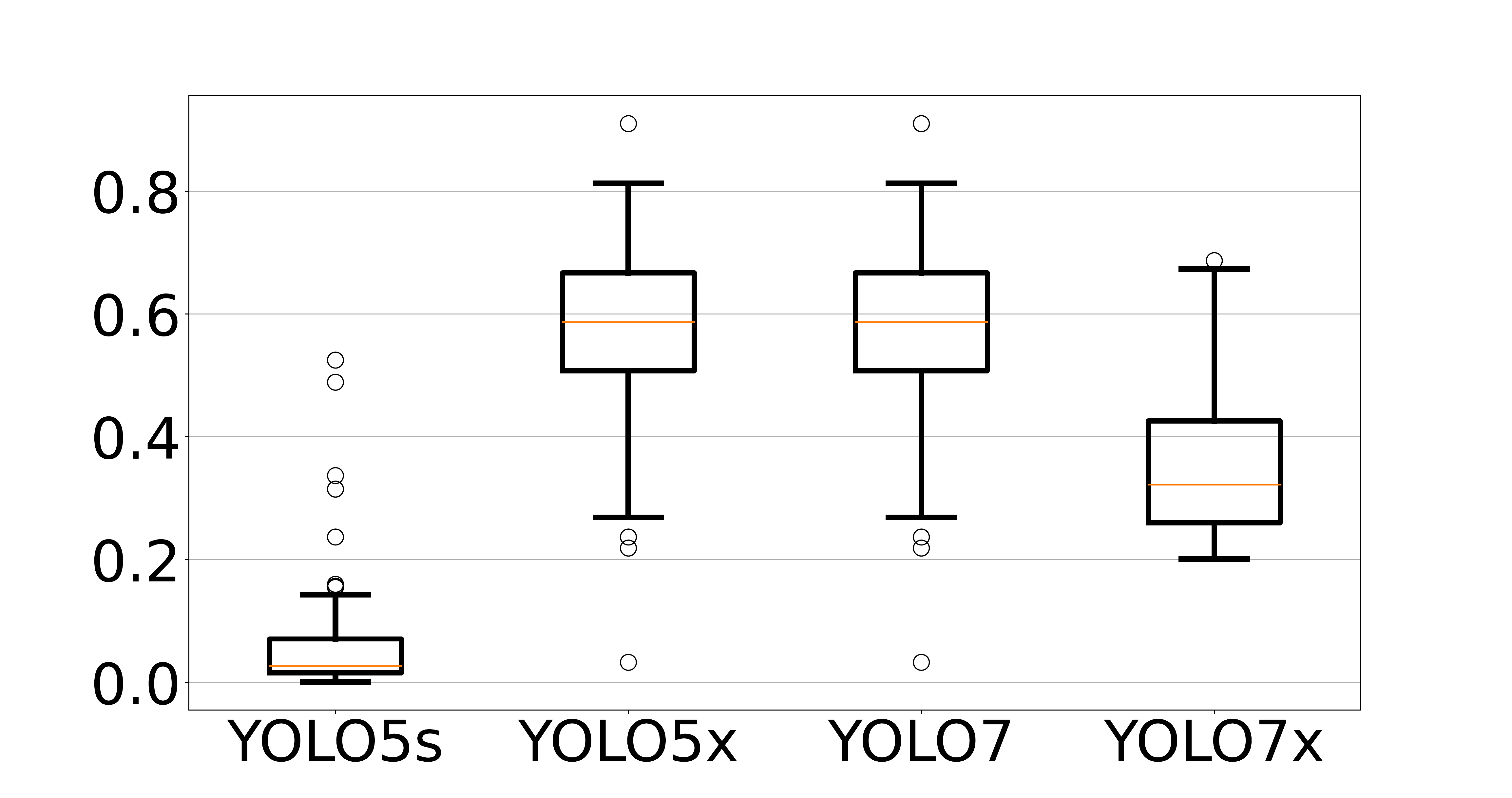}
  \caption{\textit{rain} weather condition.}
 \end{subfigure}
 \begin{subfigure}[b]{0.24\textwidth}
  \includegraphics[trim=1.8cm 0 4.5cm 0, clip, width=\textwidth,height=2.8cm]{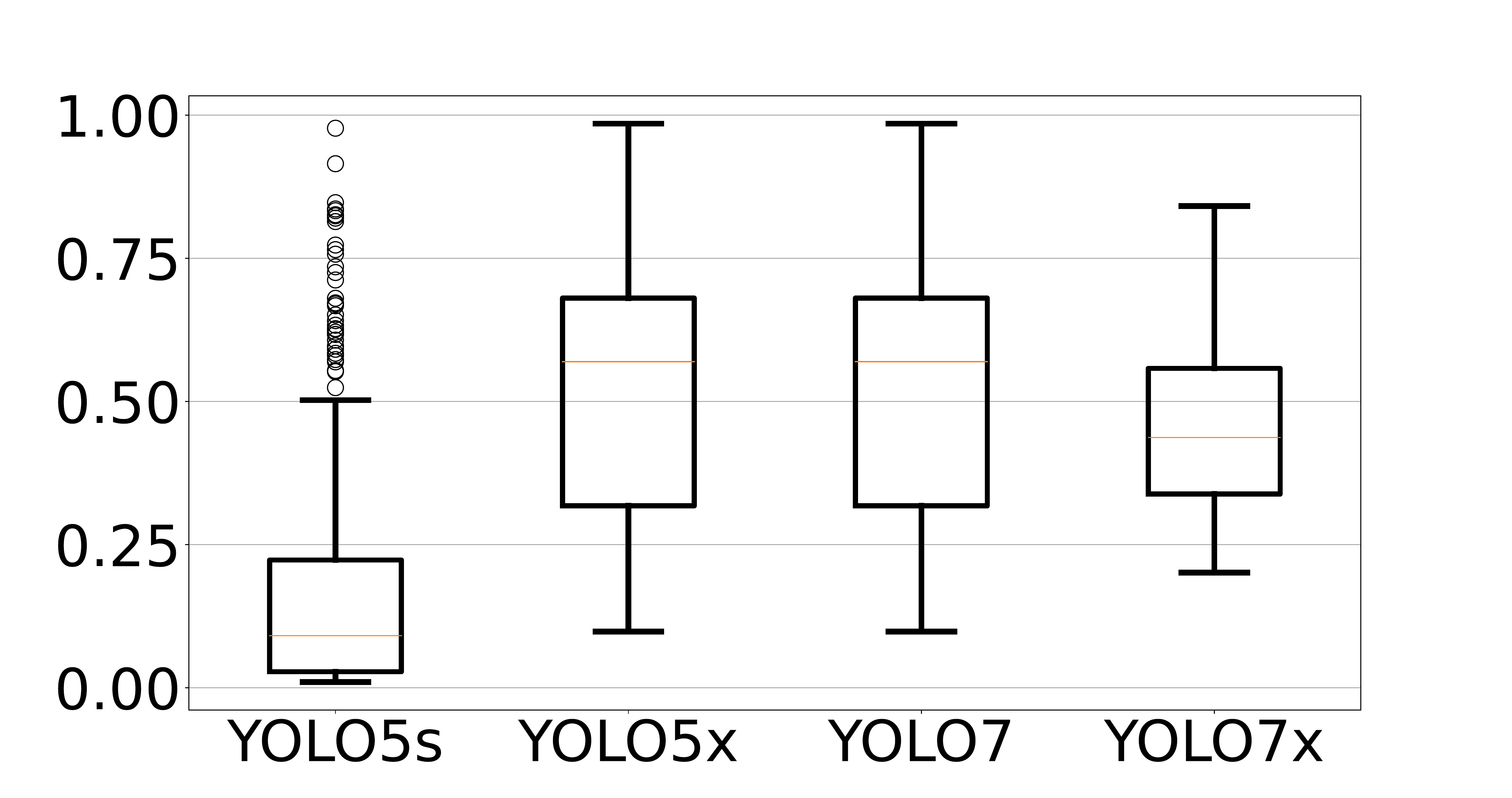}
  \caption{\textit{snow} weather condition.}
 \end{subfigure} 
 \vspace{0.2cm}
 \caption{Average Precision with IOU = 0.5  calculated for different weather conditions (\textit{sun, fog, rain and snow}), obtained for the \textit{vehicle} detection task by exploiting the four considered \textit{YOLO} methods.}
 \label{fig:car_ap}
\end{figure*}

\begin{figure}
 \centering
 \begin{subfigure}[b]{0.4\textwidth}
  \includegraphics[width=\textwidth]{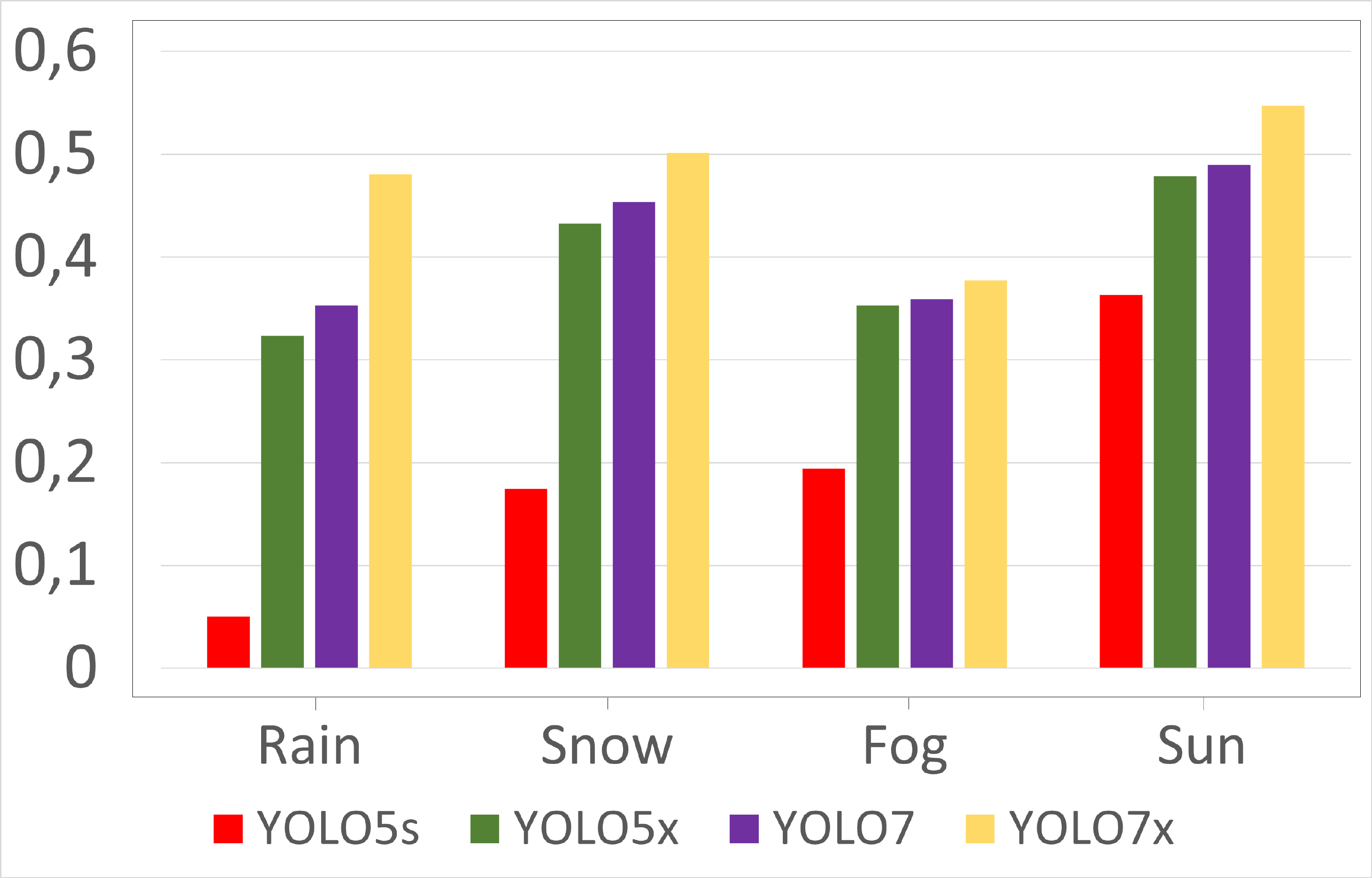}
  \caption{Results varying weather conditions.}
  \label{fig:car_sum_a}
 \end{subfigure}
 \begin{subfigure}[b]{0.44\textwidth}
  \includegraphics[width=\textwidth]{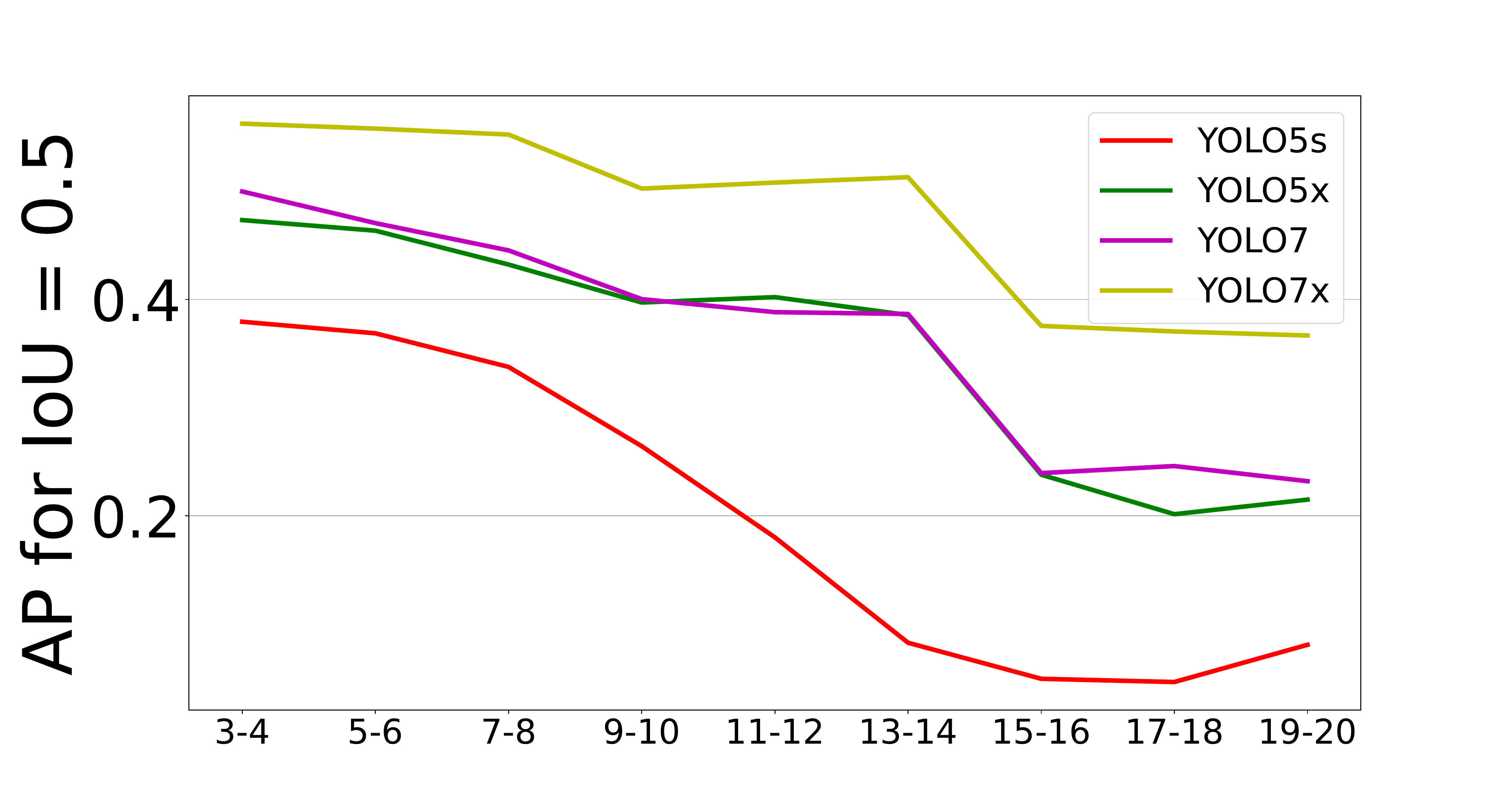}
  \caption{Results varying object densities}
  \label{fig:car_sum_b}
 \end{subfigure}
  \vspace{0.2cm}
 \caption{Summary of Average Precision with IOU = 0.5 obtained with the four \textit{YOLO}-based considered methods by varying the two main simulated factors of interest: \textit{weather condition} and \textit{density} of the objects.}
 \label{fig:car_sum}
\end{figure}

We performed two different sets of experiments --- the first related to people detection and the second to vehicle detection. We evaluated and compared the above-described detectors following the golden standard Average Precision (AP), i.e., the average precision value for recall values over 0 to 1. Specifically, we considered the MS COCO AP@[0.50], i.e., the AP computed at the single IoU threshold value of $0.50$ \cite{coco}. 
We report the results concerning people detection varying the weather conditions and the people density in Figure \ref{fig:people_ap} and Figure \ref{fig:people_sum}, respectively. On the other hand, results regarding vehicle detection varying the same two factors are depicted in Figure \ref{fig:car_ap} and Figure \ref{fig:car_sum}, respectively.

Concerning people detection, the considered models perform slightly better when the \textit{sun} weather condition is set. On the other hand, concerning the \textit{rain}, \textit{snow}, and \textit{fog} weather conditions, the detectors obtain lower APs. This is an expected outcome since, also in the real world, the detectors have to face more challenges when they are required to work in that specific conditions since the objects are more difficult to find. This trend is even more pronounced considering the car detection experiments, where some detectors particularly struggle in the \textit{rain} and \textit{fog} settings. On the other hand, the trend of both people detection and car detection exhibits performance degradation with the increasing of the objects present on the scene. Again, in this case, this behavior is expected and reflects that detecting instances is way more challenging in overcrowded scenarios.

Looking at Figure \ref{fig:people_ap}, note how in the people detection scenarios, the performance difference among the different detectors is negligible, although the \textit{YOLO7x} seems to achieve the best mean AP and the \textit{YOLO5s} exhibits the worse results. Also, considering Figure \ref{fig:people_sum_a}, we can observe how \textit{YOLO7}, \textit{YOLO7x} and \textit{YOLO5m} maintain certain robustness even in the most challenging conditions, while \textit{YOLO5s} -- besides starting with a worse detection performance even in the \textit{sun} setting -- has a decreasing trend for the other weather conditions, reaching the worst AP of around 0.19 in the \textit{fog} setting. Contrarily, the performance of the different models shows steeper differences in the car detection scenarios. In that case, the \textit{YOLO5s} completely struggles in the \textit{fog}, \textit{snow} and \textit{rain} scenarios, as shown in Figure \ref{fig:car_ap} and in Figure \ref{fig:car_sum_a}. On the other hand, \textit{YOLO7x} seems more robust to all weather conditions, except in the \textit{fog} setting, for which it exhibits moderate performances. This higher sensitivity of the detectors in the vehicle detection compared to the people scenario may be due to how the different \textit{YOLO} versions have been trained, demonstrating their major robustness to people detection -- even in very challenging weather scenarios -- than cars. This result contributes to validating our main claim that synthetic scenarios are crucial during the testing phase for finding biases or robustness breaches of largely-used detector models. Finally, by analyzing the results depicted in Figure \ref{fig:people_sum_b} and in Figure \ref{fig:car_sum_b}, we can again confirm that the performances of the considered detectors are more similar in the people detection task, while they show significant differences in detecting vehicles, especially in crowded scenarios.

\section{Conclusion}
\label{sec:conclusion}

In this work, we introduced a new synthetic dataset for \textit{people} and \textit{vehicle} detection. This collection of images is automatically annotated by interacting with a realistic simulator based on the \textit{Unity} graphical engine. This allowed us to create a vast number of different simulated scenarios leaving out humans from the annotation pipeline, in turn reducing costs and tackling the data scarcity problem affecting supervised Deep Learning models. At the same time, we kept control over some factors of interest, such as weather conditions and object densities, and we measured the performances of some state-of-the-art object detectors by varying that factors. Results showed that our simulated scenarios can be a valuable tool for measuring their performances in a controlled environment.
The presented idea has an extensive number of possible applications. People and car detection can lead to different usages, such as object counting and traffic analysis or object tracking. Furthermore, crowd simulation development is also desirable in the direction of action recognition. We also plan to enrich our simulator by introducing the possibility of viewing from multiple cameras in urban environments to create a new benchmark for multi-object tracking.

\section*{Acknowledgements}
This work was supported by: European Union funds awarded to Blees Sp. z o.o. under grant POIR.01.01.01-00-0952/20-00 “Development of a system for analysing vision data captured by public transport vehicles interior monitoring, aimed at detecting undesirable situations/behaviours and passenger counting (including their classification by age group) and the objects they carry”); EC H2020 project ``AI4media: a Centre of Excellence delivering next generation AI Research and Training at the service of Media, Society and Democracy'' under GA 951911; research project (RAU-6, 2020) and projects for young scientists of the Silesian University of Technology (Gliwice, Poland); research project INAROS (INtelligenza ARtificiale per il mOnitoraggio e Supporto agli anziani), Tuscany POR FSE CUP B53D21008060008. Publication supported under the Excellence Initiative - Research University program implemented at the Silesian University of Technology, year 2022.
This research was supported by the European Union from the European Social Fund in the framework of the project "Silesian University of Technology as a Center of Modern Education based on research and innovation” POWR.03.05.00- 00-Z098/17.
We are thankful for students participating in design of Crowd Simulator: P. Bartosz, S. Wróbel, M. Wola, A. Gluch and M. Matuszczyk.

\normalsize
\bibliography{references}

\end{document}